\documentclass[1p,12pt]{elsarticle}

\usepackage{amssymb}
\usepackage[utf8]{inputenc}
\usepackage{float}

\usepackage{amssymb}
\usepackage{booktabs}
\usepackage{multirow}
\usepackage{siunitx}
\usepackage{subfig}
\usepackage[breaklinks=true]{hyperref}
\usepackage{todonotes}
\usepackage{arydshln}
\usepackage{graphbox}
\usepackage{textpos}
\usepackage{csquotes}
\usepackage{amsmath, amsfonts}
\usepackage{dsfont}
\hypersetup{
    colorlinks=true,
    linkcolor=black,
    citecolor=black,
    urlcolor=black,
    pdfauthor=author,
    anchorcolor = black
}

\makeatletter
\def\ps@pprintTitle{   \let\@oddhead\@empty
   \let\@evenhead\@empty
   \def\@oddfoot{\reset@font\hfil\thepage\hfil}
   \let\@evenfoot\@oddfoot
}
\makeatother

\usepackage{lineno}

\begin{document}

\begin{frontmatter}

\title{Text Clustering with Large Language Model Embeddings}

\author[inst1]{Alina Petukhova\corref{cor1}}
\ead{alina.petukhova@ulusofona.pt}
\author[inst1,inst2]{João P. Matos-Carvalho}
\ead{joao.matos.carvalho@ulusofona.pt}
\author[inst1,inst2]{Nuno Fachada}
\ead{nuno.fachada@ulusofona.pt}

\cortext[cor1]{Corresponding author}
\affiliation[inst1]{organization={COPELABS, Lusófona University},
            addressline={Campo Grande, 376},
            city={Lisbon},
            postcode={1700-921},
            country={Portugal}}

\affiliation[inst2]{organization={Center of Technology and Systems (UNINOVA-CTS) and Associated Lab of Intelligent Systems (LASI)},
            city={Caparica},
            postcode={2829-516},
            country={Portugal}}

\begin{textblock*}{190mm}(-3cm,-5cm)
  \noindent \footnotesize The peer-reviewed version of this paper is
  published in the International Journal of Cognitive Computing in Engineering at
  \url{https://doi.org/10.1016/j.ijcce.2024.11.004}.
  This version is typeset by the authors and differs only in pagination and
  typographical detail.
\end{textblock*}

\vspace{-2.5cm}

\begin{abstract}
Text clustering is an important method for organising the increasing volume of digital content, aiding in the structuring and discovery of hidden patterns in uncategorised data. The effectiveness of text clustering largely depends on the selection of textual embeddings and clustering algorithms. This study argues that recent advancements in large language models (LLMs) have the potential to enhance this task. The research investigates how different textual embeddings, particularly those utilised in LLMs, and various clustering algorithms influence the clustering of text datasets. A series of experiments were conducted to evaluate the impact of embeddings on clustering results, the role of dimensionality reduction through summarisation, and the adjustment of model size. The findings indicate that LLM embeddings are superior at capturing subtleties in structured language. OpenAI's GPT-3.5 Turbo model yields better results in three out of five clustering metrics across most tested datasets. Most LLM embeddings show improvements in cluster purity and provide a more informative silhouette score, reflecting a refined structural understanding of text data compared to traditional methods. Among the more lightweight models, BERT demonstrates leading performance. Additionally, it was observed that increasing model dimensionality and employing summarisation techniques do not consistently enhance clustering efficiency, suggesting that these strategies require careful consideration for practical application. These results highlight a complex balance between the need for refined text representation and computational feasibility in text clustering applications. This study extends traditional text clustering frameworks by integrating embeddings from LLMs, offering improved methodologies and suggesting new avenues for future research in various types of textual analysis.

\end{abstract}


\end{frontmatter}

\section{Introduction}
\label{sec:introduction}

Text clustering has attracted considerable interest in text analysis due to its potential to reveal hidden structures in large volumes of unstructured textual data. With the exponential growth of digital text content generated through platforms such as social media, online news outlets, and academic publications, the ability to organise and analyse these data has become increasingly critical. Text clustering can organise large volumes of unstructured data into meaningful categories, facilitating efficient information retrieval and insightful thematic analysis across various domains, such as customer feedback, academic research, and social media content.

Text clustering serves as a preliminary step in various text analysis tasks, including topic modelling, trend analysis, and sentiment analysis. By grouping similar texts, subsequent analyses can proceed with enhanced accuracy and relevance, focusing on more homogeneous data sets with specific characteristics or themes. This process improves the precision of the analyses and ensures that the insights derived are more targeted and applicable to the context in which the data were generated.

As an analytical task, text clustering involves grouping text documents into clusters such that texts within the same cluster are more similar to each other than to those in different clusters. This process relies on the principle that text documents can be mathematically represented as vectors in a high-dimensional space, known as embeddings, where dimensions correspond to various features extracted from the documents, such as word frequency or context. Clustering algorithms use measures of proximity or resemblance to group documents that exhibit close correspondence in the feature space. This approach enables the identification of natural groupings within the data, facilitating more effective organisation and analysis of large text corpora.

The research presented in this article aims to contribute to the domain of text clustering by testing and identifying optimal combinations of embeddings---including those used in recently released large language models (LLMs)---and clustering algorithms that maximise clustering performance across various datasets. The primary objective of this paper is to determine if embeddings derived from LLMs outperform traditional embedding techniques, such as Term Frequency-Inverse Document Frequency (TF-IDF). Additionally, experiments are conducted to evaluate the impact of model size and dimensionality reduction through summarisation techniques on clustering performance.

Results indicate that LLM embeddings are highly effective at capturing the structured aspects of language, with BERT demonstrating superior performance among lightweight models. Moreover, increasing model dimensionality and employing summarisation techniques do not consistently enhance clustering efficiency, suggesting that these strategies require careful evaluation for practical application. These findings underscore the need to balance detailed text representation with computational feasibility in text clustering tasks.

This paper is organised as follows. In Section~\ref{sec:background}, advancements in textual embeddings are described, and classical text clustering algorithms used in this domain are briefly mentioned. Section~\ref{sec:methods} outlines the main steps and components of this study, including dataset selection, data preprocessing, embeddings, and clustering algorithm configurations used to assess clustering quality. Section~\ref{sec:results_discussion} presents the results of our study and provides a discussion of these findings. Limitations encountered during the study, along with recommendations for overcoming them, are acknowledged in Section~\ref{sec:limitations}. Finally, Section~\ref{sec:conclusion} synthesises the main conclusions of this research and suggests future developments.

\section{Background}
\label{sec:background}

\subsection{Text Embeddings}
\label{subsec:textembeddings}

The field of text representation in natural language processing (NLP) has undergone an impressive transformation over the past few decades. From simple representations to highly sophisticated embeddings, advancements in this domain have significantly improved the ability of machines to process and understand human language with increasing accuracy.

One of the earliest methods of text representation that laid the groundwork for subsequent advances was Term Frequency-Inverse Document Frequency (TF-IDF). This method quantifies the importance of a word within a document relative to a corpus by accounting for term frequency and inverse document frequency~\cite{Salton1983}. While TF-IDF effectively highlights the relevance of words, it treats each term as independent and fails to capture word context and semantic meaning~\cite{Ramos2003}.

Word embeddings, such as those produced by Word2Vec~\cite{Mikolov2013} and GloVe~\cite{Pennington2014}, marked a significant advancement by generating dense vector representations of words based on their contexts. These models leveraged surrounding words and large corpora to learn word relationships, successfully capturing a range of semantic and syntactic similarities. Despite their effectiveness in capturing semantic regularities, these models still provided a single static vector per word, which posed limitations in handling polysemous words---words with multiple meanings.

The arrival of BERT (Bidirectional Encoder Representations from Transformers) initiated a new phase of embedding sophistication~\cite{Devlin2018}. To generate contextual embeddings, BERT employs a bidirectional transformer architecture, pre-trained on a massive corpus. This allows for a deeper understanding of word relationships by considering the full context of a word in a sentence in both directions. BERT revolutionised tasks like text clustering by providing richer semantic representations.

Today, LLMs like OpenAI's GPT are at the forefront of generating state-of-the-art embeddings~\cite{Brown2020}. LLMs extend the capabilities of previous models by providing an unprecedented depth and breadth of knowledge encoded in word and sentence-level embeddings. These models are trained on extensive datasets to capture a broad spectrum of human language variations and generate embeddings that reflect a comprehensive understanding of contexts and concepts.

The progression from TF-IDF to sophisticated LLM embeddings represents a significant advancement towards more contextually aware text representation in NLP. This evolution continues to propel the field forward, expanding the possibilities for applications such as text clustering, sentiment analysis, and beyond. In the context of text clustering methodologies, there exists a considerable research gap that underscores the need for a comprehensive evaluation of LLM embeddings against traditional techniques such as TF-IDF.

\subsection{Text Clustering Algorithms}
\label{subsec:clusteringalgorithms}

Text clustering involves grouping a set of texts such that texts in the same group (referred to as a cluster) are more similar to each other than to those in different clusters. This section provides an overview of classic clustering algorithms widely used for clustering textual data.

$K$-means is perhaps the most well-known and commonly used clustering algorithm due to its simplicity and efficiency. It partitions the dataset into $k$ clusters by minimising the within-cluster sum of squares, i.e., variance. Each cluster is represented by the mean of its points, known as the centroid~\cite{MacQueen1967}. $K$-means is particularly effective for large datasets but depends heavily on the initialisation of centroids and the value of $k$, which must be known a priori.

Agglomerative hierarchical clustering (AHC) builds nested clusters by merging them successively. This bottom-up approach starts with each text as a separate cluster and combines clusters based on a linkage criterion, such as the minimum or maximum distance between cluster pairs~\cite{Ward1963}. AHC is versatile and allows for discovering hierarchies in data, but it can be computationally expensive for large datasets.

Spectral clustering techniques use the eigenvalues of a similarity matrix to reduce dimensionality before applying a clustering algorithm such as $k$-means. It is particularly adept at identifying clusters that are not necessarily spherical, as $k$-means assumes. Spectral clustering can also handle noise and outliers effectively~\cite{Ng2002}. However, its computational cost can be high due to the eigenvalue decomposition involved.

Fuzzy $c$-means (FuzzyCM) is a clustering method that allows data points to belong to more than one cluster. This method minimises the objective function with respect to membership and centroid position, providing soft clustering and handling overlapping clusters~\cite{Bezdek1981}. FuzzyCM is useful when the boundaries between clusters are not clearly defined, although it is computationally more intensive than $k$-means.

In addition to these classic approaches, recent years have seen the rise of alternative methods for text clustering that leverage the strengths of modern embeddings and consider the unique properties of textual data. These include using deep learning models, particularly those based on autoencoders, to learn meaningful low-dimensional representations ideal for clustering~\cite{Xie2016,KAMAL2022}.

Ensemble clustering approaches, where multiple clustering algorithms are combined to improve the robustness and quality of the results, have also been gaining attention. These methods benefit from the diversity of the individual algorithms and can sometimes overcome the limitations of any single method~\cite{Strehl2002}.

The field of text clustering is rapidly expanding, and keeping up with the latest findings is crucial for advancing the state of the art. Recent papers, such as those exploring the use of transformer-based embeddings for clustering~\cite{Pugachev2021} and integrating external knowledge bases into the clustering process~\cite{zhang2019integrating}, represent just the tip of the iceberg in this area of research. Additionally, very recent work by Keraghel et al.~\cite{Keraghel2024} offers a preliminary discussion on the potential of using LLM embeddings for text clustering. The authors analysed embeddings from BLOOMZ, Mistral, Llama-2, and OpenAI using five clustering algorithms. This paper extends that work by focusing on different datasets, testing additional LLM embeddings, and comparing results using classical TF-IDF as a baseline. Furthermore, this study experiments with summarisation as a dimensionality reduction technique and evaluates the impact of model size on clustering results.

\section{Methods}
\label{sec:methods}

This research evaluates the effectiveness of various embedding representations in enhancing the performance of text clustering algorithms, aiming to identify the most informative embeddings for clustering tasks. To achieve this, we systematically experimented with multiple datasets, embeddings, and clustering methods, and performed an in-depth analysis of clustering results using different evaluation metrics. The following steps were undertaken during the course of this study:

\begin{enumerate}
\item Selection of datasets, ensuring the robustness of our findings across different types of textual data.
\item Preprocessing of datasets, including the removal of miscellaneous characters such as emails and HTML tags.
\item Utilisation of various embedding computations, including LLM-related ones, to retrieve numerical text representations.
\item Application of several clustering algorithms commonly used in text clustering.
\item Comparison of clustering results using different external and internal validation metrics.
\end{enumerate}

The following subsections provide a comprehensive description of each of these steps, further detailing the employed methodologies.

\subsection{Datasets}
\label{subsec:datasets}

We selected five datasets to cover a variety of text clustering challenges. Table~\ref{tab:datasets} shows these datasets and their characteristics. The CSTR abstracts dataset~\cite{yamagishi2019vctk} is a corpus of 299 scientific abstracts from the Centre for Speech Technology Research. The homogeneous and domain-specific nature of the CSTR dataset allowed us to investigate the effectiveness of clustering techniques in discerning fine-grained topic distinctions in scholarly text related to categories such as artificial intelligence, theory, systems, and robotics.

The SyskillWebert dataset~\cite{SyskillWebert_dataset}, which includes user ratings for web pages, enabled exploratory clustering to analyse information used in recommendation systems. The 20Newsgroups dataset~\cite{Newsgroups20} is a famous collection of approximately 19,000 news documents partitioned across 20 different classes. With its broad assortment of topics and noisy, unstructured text, this dataset provided a realistic scenario for evaluating the robustness of clustering algorithms under less-than-ideal conditions.

To complement these, we added the MN-DS dataset~\cite{data8050074}, a diverse compilation of multimedia news articles, which provides the opportunity to explore the effectiveness of clustering algorithms in handling multi-categorical data. The dataset is organised hierarchically in two levels; therefore, experiments were performed independently for each level. Additionally, we included the Reuters dataset~\cite{lewis2004reuters}, a benchmark dataset widely used in text mining and information retrieval research, which contains stories from the Reuters news agency. To enhance the comprehensiveness of our evaluation, we selected documents assigned to single classes, reducing the dataset from 10,369 to 8,654 text articles. This allowed us to assess the clustering algorithms' ability to accurately group similar documents according to their predefined categories.

\begin{table}[!ht]
    \caption{Tested datasets and their characteristics. `Size' represents the number of documents in the dataset, while `No. classes' is the number of categories.}
    \label{tab:datasets}
    \begin{center}
    {\small
    \begin{tabular}{lrrc}
    \toprule
    Dataset & Size & No. classes & Ref. \\
    \midrule
    CSTR & 299 & 4 & \cite{yamagishi2019vctk} \\
    SyskillWebert & 333 & 4 & \cite{SyskillWebert_dataset} \\
    20Newsgroups dataset & 18,846 & 20 & \cite{Newsgroups20} \\
    MN-DS dataset level 1 & 10,917 & 17 & \cite{data8050074} \\
    MN-DS dataset level 2 & 10,917 & 109 & \cite{data8050074} \\
    Reuters & 8,654 & 65 & \cite{lewis2004reuters} \\
    \bottomrule
    \end{tabular}
    }
    \end{center}
\end{table}

\subsection{Preprocessing}
\label{subsec:preprocessing}

The motivation for preprocessing text data is to minimise noise and highlight key patterns, thereby improving the efficiency and accuracy of clustering algorithms~\cite{textcl}.

For all datasets, a series of preprocessing steps were taken to ensure the quality and uniformity of the input data. The initial step involved removing miscellaneous items such as irrelevant metadata, HTML tags, and any extraneous content that might skew the analysis. Continuing with this methodology, we systematically eliminated invalid non-Latin characters from the dataset. These characters could arise from multilingual data sources or artefacts of data collection, such as encoding errors. Given that the employed workflow is optimised for Latin-based languages, retaining such characters could increase the dimensionality of the feature space, thereby undermining clustering performance, both in terms of computational efficiency and interpretability of the results.

\subsection{Text Embeddings}
\label{subsec:embedding}

For textual data representation, we compared classical embeddings~\cite{Devlin2018,linkTFIDF} and state-of-the-art LLM embeddings. Traditional TF-IDF vectors served as a baseline, providing a sparse but interpretable representation based on word importance within the analysed dataset. BERT embeddings were extracted from the BERT model, which was trained as a transformer-based bidirectional encoder on BookCorpus~\cite{Zhu_2015_ICCV} and English Wikipedia~\cite{przybyla_2022_6539054}. These embeddings were utilised to achieve deep contextual understanding, capturing the semantic variations across the corpus.

We also employed ``text-embedding-ada-002''~\cite{Greene2023Embedding} embeddings, as they demonstrated the best results among OpenAI's embeddings on larger datasets for tasks such as text search, code search, and sentence similarity. Additionally, other LLM embeddings were extracted for Falcon~\cite{almazrouei2023falcon} and LLaMA-2-chat~\cite{touvron2023llama} models, included for their respective advancements in performance and efficiency. Falcon embeddings were trained on a hybrid corpus consisting of both text documents and code, while the LLaMA-2-chat embeddings---built on the foundation of the LLaMA-2 model using an optimised auto-regressive transformer---underwent targeted fine-tuning for dialogue and question-and-answer tasks.

BERT, Falcon, and LLaMA-2-chat embeddings were obtained from Hugging Face's transformers library, a platform providing state-of-the-art models in accessible pipelines~\cite{huggingface}. Table~\ref{tab:emb} describes the exact embeddings and configurations used.

\begin{table}[H]
    \caption{Tested embeddings and their configuration. For TF-IDF, parameters $\mathsf{min\_df}$ and $\mathsf{max\_df}$ are used to exclude terms that have a document frequency strictly lower than or higher than the specified thresholds, respectively, during the vocabulary creation process; if real-valued, these parameters represent a proportion of documents; if integer-valued, they represent absolute document counts. In turn, $\mathsf{max\_features}$ limits the vocabulary to only the top $\mathsf{max\_features}$ terms, ordered by term frequency across the corpus. }
    \label{tab:emb}
    \begin{center}
    {\small
    \begin{tabular}{lll}
    \toprule
    Embeddings & Configuration & \\
    \midrule
    TF-IDF &
      $\mathsf{min\_df}=5$, $\mathsf{max\_df}=0.95$, $\mathsf{max\_features}=8000$ \\
    BERT &
      huggingface.co/sentence-transformers/all-mpnet-base-v2 \\
    OpenAI & huggingface.co/Xenova/text-embedding-ada-002
        \\
    Falcon &
      huggingface.co/tiiuae/falcon-7b\\
    LLaMA-2 &
      huggingface.co/meta-llama/Llama-2-7b-chat-hf \\
    \bottomrule
    \end{tabular}
    }
    \end{center}
\end{table}

\subsection{Clustering Algorithms}
\label{subsec:algorithms}

The selected clustering algorithms address the diverse nature of text data, which often contains complex patterns requiring robust methods for effective grouping. Standard $k$-means clustering was used for its simplicity and efficiency in dealing with large datasets. $K$-means++ was chosen as an enhanced variant of $k$-means, with careful initialisation to improve convergence and cluster quality~\cite{arthur2007k}. AHC was utilised for its ability to reveal nested structures within the data. Compared to $k$-means, which assigns each data point to a single cluster, FuzzyCM provides a probabilistic membership approach, accommodating polysemy and subtle semantic differences typical in text data. Finally, Spectral clustering was selected for its effectiveness in identifying clusters based on their data-induced graph structure, and it is particularly adept at discovering clusters with non-convex shapes.

To map the newly formed clusters to the original labels of the dataset, we calculated the Euclidean distance between the centroid of each derived cluster and the centroid of each ground truth cluster, assigning the closest ones. The Euclidean distance is a preferable metric for this process because it accurately reflects distances in multidimensional space, ensuring precise mapping. This method allows the use of external evaluation metrics such as the $F1$-score. In our work, we did not apply Euclidean distance at the word level, which can be problematic. Instead, the process computes an aggregation for each identified cluster, enabling a more reliable association with the existing labels.

The selected algorithms and their respective parameters are listed in Table~\ref{tab:algs}. The \textit{scikit-learn} library~\cite{scikit-learn} provided the implementations for all algorithms except for FuzzyCM, which was utilised from the \textit{scikit-fuzzy} package~\cite{sfuzzy}.

\begin{table}[H]
    \caption{Tested clustering algorithms and respective parameters. For $k$-means and $k$-means++, $\mathsf{init}$~refers to the method used for the initial selection of cluster centroids, $\mathsf{n_{init}}$ specifies the number of times the $k$-means algorithm is run with different centroid seeds, while $\mathsf{seed}$ defines the random number for centroid initialisation. For AHC, $\mathsf{metric}$ defines the metric used to compute the linkage, while $\mathsf{linkage}$ specifies the linkage criterion that determines the distance measurement used between sets of observations. For FuzzyCM, $\mathsf{init}$ is the initial fuzzy $c$-partitioned matrix---if set to $None$, the matrix will be randomly initialised---$\mathsf{m}$~is the degree of fuzziness, $\mathsf{error}$ defines the stopping criterion, and $\mathsf{maxiter}$ specifies the maximum number of iterations allowed. For Spectral clustering, $\mathsf{assign\_labels}$ determines the strategy used to assign labels in the embedding space and $\mathsf{seed}$ is a pseudo-random number seed for initialising the locally optimal block preconditioned conjugate gradient eigenvectors decomposition. For all algorithms, the number of clusters was set to match the number of classes present in the dataset.}
    \label{tab:algs}
    \begin{center}
    {\small
    \begin{tabular}{lll}
    \toprule
    Algorithm & Parameters & \\
    \midrule
    $k$-means &
      \multicolumn{2}{l}{
    $\mathsf{init}=\text{random}$, $\mathsf{n_{init}}=10$, $\mathsf{seed}=0$} \\
    $k$-means++ &
      \multicolumn{2}{l}{
      $\mathsf{init}=\text{$k$-means++}$, $\mathsf{n_{init}}=1$, $\mathsf{seed}=0$} \\
    AHC &
      \multicolumn{2}{l}{
        $\mathsf{metric}=\text{euclidean}$,  $\mathsf{linkage}=\text{ward}$} \\
    FuzzyCM &
      \multicolumn{2}{l}{
        $\mathsf{init}=\text{None}$,  $\mathsf{m}=2$,
        $\mathsf{error}=0.005$,
        $\mathsf{maxiter}=1000$
        } \\
    Spectral &
        \multicolumn{2}{l}{
        $\mathsf{assign\_labels}=\text{discretize}$,
        $\mathsf{seed}=10$} \\
    \bottomrule
    \end{tabular}
    }
    \end{center}
\end{table}

\subsection{Evaluation Metrics}
\label{subsec:metrics}

To comprehensively evaluate the quality of different embeddings and algorithm combinations, we used a diverse set of metrics. For external validation, since the original labels were available, we used the weighted $F1$-score (F1S)~\cite{evaluationmetrics}, the Adjusted Rand Index (ARI)~\cite{SteinleyDouglas2004}, and the Homogeneity score (HS)~\cite{Rosenberg2007}. The $F1$-score was computed to balance precision and recall in the presence of class imbalance. ARI was used to assess clustering outcomes while correcting for chance grouping, and HS was used to evaluate the degree to which each cluster is composed of data points primarily from one class. For internal validation, we employed the Silhouette Score (SS)~\cite{Rousseeuw1987} and the Calinski-Harabasz Index (CHI)~\cite{Caliński1974}, evaluating cluster coherence and separation without requiring ground truth. This multifaceted approach ensures a robust assessment, capturing both the alignment with known labels and the intrinsic structure of the generated clusters. These metrics collectively provide a balanced view of performance, accounting for datasets with varying characteristics and sizes. The metrics and their corresponding formulas are presented in Table~\ref{tab:metrics}.

\begin{table}[H]
    \caption{Metrics used to assess clustering results, their type (external or internal), and their respective formulas. For the $F1$-score (F1S), $C$ represents the number of classes, $w_{i}$ is the weight assigned to the $i$-th class, which is typically the proportion of that class within the dataset, and $F_{1,i}$ represents the $F1$-score computed for the $i$-th class. For the Adjusted Rand Index (ARI), $RI$ stands for Rand Index, $Expected\_RI$ refers to the expected value of the Rand Index under random label assignment (calculated using the contingency table marginals), and $Max\_RI$ is the maximum possible value of the Rand Index. For the Homogeneity Score (HS), $H(C|K)$ is the conditional entropy of the class distribution given the predicted cluster assignments, and $H(C)$ is the entropy of the class distribution. For the Silhouette Score (SS), $N$ represents the total number of data points in the dataset, and $s(i)$ is the silhouette score for a single data point $i$, defined as $s(i) = \frac{b(i) - a(i)}{\max\{a(i), b(i)\}}$, where $a(i)$ is the average distance from the $i$-th data point to other points in the same cluster, and $b(i)$ is the minimum mean distance from the $i$-th data point to points in a different cluster, minimised over all clusters. For the Calinski-Harabasz Index (CHI), $Tr(B_{k})$ is the trace of the between-group dispersion matrix that measures the between-cluster dispersion, $Tr(W_{k})$ is the trace of the within-cluster dispersion matrix, which quantifies the within-cluster dispersion, $N$ refers to the number of data points, and $k$ indicates the number of clusters. The optimal value for all these metrics is the maximum.}
    \label{tab:metrics}

    \begin{center}
    {\small
    \begin{tabular}{llll}
    \toprule
    Metric & Type & Formula \\
    \midrule
    F1S & External &
      \(\displaystyle \sum_{i=1}^{C} w_i F_{1,i}\) \\
    ARI & External &
      \(\displaystyle \frac{RI - Expected\_RI}{Max\_RI - Expected\_RI} \) \\
    HS & External &
      \( 1 - \frac{H(C|K)}{H(C)} \) \\
    SS & Internal &
      \( \frac{1}{N} \sum_{i=1}^{N} s(i) \) \\
    CHI & Internal &
      \( \frac{\text{Tr}(B_k)}{\text{Tr}(W_k)} \times \frac{N - k}{k - 1} \)\\
    \bottomrule
    \end{tabular}
    }
    \end{center}
\end{table}

\subsection{Additional Experiments}
\label{subsec:experiment}

This section describes additional experiments in which we performed text summarisation (\ref{subsec:summarisation}) and tested LLM embeddings obtained from larger models (i.e., models with more parameters) prior to clustering (\ref{subsec:dim}). The purpose of these experiments is to investigate whether such representations can improve the discriminability of features within text clusters.

\subsubsection{Summarisation}
\label{subsec:summarisation}

This experiment aims to evaluate summarisation as a tool for dimensionality reduction in text clustering by creating compact representations of the texts that encapsulate their semantic core without losing context. These experiments are hypothesised to streamline the clustering process, potentially leading to more coherent and interpretable clusters, even in large and complex datasets. This involves adding a summarisation step after preprocessing and before the retrieval of embeddings. The models used in summarisation are described in Table~\ref{tab:summodels}. As an alternative approach to LLM-based models, we used the BERT-large-uncased summarisation model~\cite{Devlin2018} implemented by the BERT summariser~\cite{miller2019leveraging} to assess potential improvements in clustering achieved by utilising a lower-dimensionality model.

\begin{table}[H]
    \caption{Overview of text summarisation models used in this study, each employing a transformer-based architecture. `Max Tokens' is the maximum input sequence length. BERT is a bidirectional encoder-only model trained on BookCorpus, a dataset consisting of 11,038 unpublished books and English Wikipedia (excluding lists, tables, and headers); OpenAI is a GPT-3-based decoder-only model built with multi-head attention blocks, trained on an extended dataset with reinforcement learning from human feedback; Falcon is a causal decoder-only model with a multi-query attention mechanism trained on 1,500B tokens of RefinedWeb enhanced with curated corpora; LLaMA-2-chat is a LLaMA-2-based auto-regressive decoder-only model trained on a 1,000B token dataset collected by Meta, enriched with supervised fine-tuning, reinforcement learning from human feedback, and the Ghost Attention mechanism.}
    \label{tab:summodels}
    \begin{center}
    {\small
    \begin{tabular}{p{1.0in}p{1.8in}r}
    \toprule
    Embeddings & Summarisation Model & Max Tokens  \\
    \midrule
    BERT & bert-large-uncased~\cite{Devlin2018} & 512 \\
    OpenAI & gpt-3.5-turbo~\cite{gpt35turbo} & 4096 \\
    Falcon & falcon-7b~\cite{falcon7b,almazrouei2023falcon} & 2048 \\
    LLaMA-2-chat & Llama-2-7b-chat-hf~\cite{Llama27bchathf,touvron2023llama} & 4096 \\
    \bottomrule
    \end{tabular}
    }
    \end{center}
\end{table}

For the LLaMA-2 and Falcon models, we used the Hugging Face transformers library with the parameters described in Table~\ref{tab:sumparams}.

\begin{table}[H]
    \caption{Parameters used for summarisation with LLaMA-2 and Falcon models: $\mathsf{temperature}$ represents the value to modulate probabilities of the next token, $\mathsf{max\_length}$ defines the maximum length of the sequence to be generated, $\mathsf{do\_sample}$ is a parameter that determines whether to use sampling rather than greedy decoding, $\mathsf{top\_k}$ restricts the selection to the top $k$ tokens with the highest probabilities during top-$k$ filtering, and $\mathsf{num\_return\_sequences}$ is the number of independently computed returned sequences for each element in the batch.}
    \label{tab:sumparams}
    \begin{center}
    {\small
    \begin{tabular}{ll}
    \toprule
    Parameter name & Value  \\
    \midrule
    $\mathsf{temperature}$ & 0 \\
    $\mathsf{max\_length}$ & 800 \\
    $\mathsf{do\_sample}$ & True \\
    $\mathsf{top\_k}$ & 10 \\
    $\mathsf{num\_return\_sequences}$ & 1 \\
    \bottomrule
    \end{tabular}
    }
    \end{center}
\end{table}

The following zero-shot prompt was used for generating the summarised text with LLMs:

\begin{displayquote}
\textit{Write a concise summary of the text. Return your responses with maximum 5 sentences that cover the key points of the text. \\
\{text\} \\
SUMMARY:}
\end{displayquote}

\subsubsection{Increasing Model Dimension}
\label{subsec:dim}

The original publications on LLMs underline the performance increase with larger model sizes in tasks such as common sense reasoning, question answering, and code tasks~\cite{almazrouei2023falcon,touvron2023llama}. This experiment evaluates embeddings from various LLM sizes to analyse the impact of higher-dimensional models on the performance of clustering algorithms, aiming to determine if they enhance cluster cohesion and separation. For this purpose, we used embeddings obtained from models presented in Table~\ref{tab:embeddings-highdim}.

To visualise the different embeddings and capture their intrinsic structures, Principal Component Analysis (PCA) and $t$-Distributed Stochastic Neighbor Embedding ($t$-SNE)~\cite{van2008visualizing} were employed. Initially, PCA was applied for preliminary dimensionality reduction while preserving variance. Subsequently, $t$-SNE was used to project the data into a lower-dimensional space, emphasising local disparities between embeddings. This sequential application of PCA and $t$-SNE allows us to capture both global and local structures within the embeddings, providing a richer visualisation than using PCA alone.

\begin{table}[!ht]
    \caption{Embeddings and corresponding models used in the dimensionality experiments. Here, `Size' represents the number of parameters denoted in billions (bp), and `Tokens' refers to the embedding token size in trillions (trl).}
    \label{tab:embeddings-highdim}
    \begin{center}
    \begin{tabular}{lp{2.6in}p{0.5in}p{0.5in}}
    \toprule
    Model & Model Reference & Size (bp) & Tokens (trl) \\
    \midrule
    Falcon-7b & huggingface.co/tiiuae/falcon-7b & 7 & 1.5 \\
    Falcon-40b & huggingface.co/tiiuae/falcon-40b & 40 & 1 \\
    LLaMA-2-7b & huggingface.co/meta-llama/Llama-2-7b-chat-hf & 7 & 2 \\
    LLaMA-2-13b & huggingface.co/meta-llama/Llama-2-13b-chat-hf & 13 & 2 \\
    \bottomrule
    \end{tabular}
    \end{center}
\end{table}

\section{Results and Discussion}
\label{sec:results_discussion}

Table~\ref{figres1} presents the clustering metrics for the ``best'' algorithm for the tested combinations of dataset, embedding, and clustering algorithm. By ``best'', we mean the algorithm with the highest $F1$-score value. The complete results are available as supplementary material\footnote{\url{https://doi.org/10.5281/zenodo.10844657}}.

Results demonstrate that OpenAI embeddings generally yield superior clustering performance on structured, formal texts compared to other methods based on most metrics (column `Total' in Table~\ref{figres1}). The combination of the $k$-means algorithm and OpenAI's embeddings yielded the highest values of ARI, $F1$-score, and HS in most experiments. This may be attributed to OpenAI's embeddings being trained on a diverse array of Internet text, rendering them highly effective at capturing the diversity of language structures.

Low values of SS and CHI for the same algorithm could indicate that, while clusters are homogeneous and aligned closely with the ground truth labels (suggesting good class separation and cluster purity), they may not be well-separated or compact as evaluated in a geometric space. This discrepancy can arise in cases where data has a high-dimensional or complex structure that external measures such as ARI, F1S, and HS capture effectively. However, when projecting into a lower-dimensional space for SS and CHI, the clusters appear to overlap or vary widely in size, leading to a lower score for these spatial coherence metrics. This highlights a challenge in text clustering: achieving both high cluster purity alongside tight cluster geometry.

Interestingly, for the case of the Reuters dataset (DS6), we observed that FuzzyCM consistently outperformed other clustering algorithms when paired with different embeddings. The potential reason behind this result may be related to the flexibility of fuzzy clustering, which allows data points to belong to multiple clusters with varying degrees of membership, making it particularly suitable for datasets with a large number of classes such as the Reuters dataset.

In the domain of open-source models, namely Falcon, LLaMA-2, and BERT, the latter emerged as the frontrunner. Given that BERT is designed to understand context and potentially due to the model's lower dimensionality, these embeddings demonstrate good effectiveness in text clustering. In the comparative analysis of open-source LLM embeddings, Falcon-7b outperformed LLaMA-2-7b across most datasets, demonstrating improved cluster quality and distinctiveness. This superiority may be attributed to Falcon-7b embeddings' ability to better capture salient linguistic features and semantic relationships within the texts since these embeddings were trained on a mixed corpus of text and code, as opposed to the LLaMA-2 embeddings, which are specialised for dialogues and Q\&A contexts. Additionally, the CHI metric---measuring the dispersion ratio between and within clusters---is higher for Falcon-7b embeddings, suggesting that clusters are well-separated and dense.

Experiments with the MN-DS dataset, featuring a hierarchical label structure, indicate that clustering at a higher, more abstract label level (17 classes) produces better class separation. However, the $F1$-score is higher when clustering is assessed at the more specific level (109 classes). This likely indicates better precision and recall for individual classes, while lower values for other metrics reflect the increased difficulty in maintaining overall clustering quality and cohesion with a higher number of classes. These results indicate that clustering at higher levels can produce more cohesive and interpretable clusters that align with natural categorical divisions, though at the expense of extracting less specific information for each document.

\begin{table}[!htbp]
\small
\caption{Results of text clustering for the best-performing clustering algorithms for each combination of dataset and embedding. The best algorithm was determined by choosing the algorithm with the highest $F1$-score value. DS1 represents the CSTR dataset, DS2 is the SyskillWebert dataset, DS3 is the 20newsgroups dataset, DS4 is the MN-DS dataset for level 1 labels, DS5 is the MN-DS dataset for level 2 labels, and DS6 is the Reuters dataset. `Total' represents the number of metrics per given embeddings/algorithm combination that outperform other combinations.}
\label{figres1}

\begin{tabular}{p{0.45in}p{0.7in}p{0.85in}wr{0.7cm}wr{0.7cm}wr{0.7cm}wr{0.9cm}wr{0.65cm}wr{0.7cm}}
\toprule
Dataset & Embed. & Best Alg. & F1S & ARI & HS & SS & CHI & Total \\
\midrule
DS1 & TF-IDF & $k$-means & 0.67 & 0.38 & 0.46 & 0.016 & 4 & 0/5 \\
 & BERT & Spectral & \textbf{0.85} & \textbf{0.60}  & 0.63 & \textbf{0.118} & 25 & 3/5 \\
 & OpenAI & $k$-means & 0.84 & 0.59 & \textbf{0.64} & 0.066 & 13 & 1/5 \\
 & LLaMA-2 & $k$-means & 0.41 & 0.09 & 0.17 & 0.112 & \textbf{49} & 1/5 \\
 & Falcon & $k$-means & 0.74 & 0.39 & 0.48 & 0.111 & 34 & 0/5 \\
\midrule
DS2 & TF-IDF & Spectral & 0.82 & 0.63 & 0.58 & 0.028 & 8 & 0/5 \\
 & BERT & AHC & 0.74 & 0.58 & 0.53 & 0.152 & 37 & 0/5 \\
 & OpenAI & AHC & \textbf{0.90} & \textbf{0.79} & \textbf{0.75} & 0.070 & 19 & 3/5 \\
 & LLaMA-2 & $k$-means & 0.51 & 0.21 & 0.25 & 0.137 & 69 & 0/5 \\
 & Falcon & $k$-means++ & 0.45 & 0.26 & 0.30 & \textbf{0.170} & \textbf{85} & 2/5 \\
\midrule
DS3 & TF-IDF & Spectral & 0.35 & 0.13 & 0.28 & -0.002 & 37 & 0/5 \\
 & BERT & $k$-means & 0.43 & 0.25 & 0.44 & 0.048 & 412 & 0/5 \\
 & OpenAI & $k$-means & \textbf{0.69} & \textbf{0.52} & \textbf{0.66} & 0.035 & 213 & 3/5 \\
 & LLaMA-2 & AHC & 0.17 & 0.11 & 0.26 & 0.025 & 264 & 0/5 \\
 & Falcon & $k$-means & 0.26 & 0.15 & 0.30 & \textbf{0.071} & \textbf{1120} & 2/5 \\
\midrule
DS4 & TF-IDF & $k$-means & 0.29 & 0.13 & 0.48 & 0.034 & 17 & 0/5 \\
 & BERT & $k$-means & 0.35 & 0.24 & 0.55 & \textbf{0.072} & 61 & 1/5 \\
 & OpenAI & $k$-means & \textbf{0.38} & \textbf{0.26} & \textbf{0.58} & 0.053 & 42 & 3/5 \\
 & LLaMA-2 & $k$-means & 0.21 & 0.11 & 0.40 & 0.053 & 88 & 0/5 \\
 & Falcon & $k$-means++ & 0.27 & 0.16 & 0.48 & 0.071 & \textbf{92} & 1/5 \\
\midrule
DS5 & TF-IDF & AHC & 0.31 & 0.09 & 0.29 & 0.010 & 37 & 0/5 \\
 & BERT & $k$-means++ & 0.43 & \textbf{0.27} & \textbf{0.42} & 0.060 & 178 & 2/5 \\
 & OpenAI & Spectral & \textbf{0.45} & 0.25 & 0.41 & 0.036 & 120 & 1/5 \\
 & LLaMA-2 & AHC & 0.23 & 0.10 & 0.23 & 0.031 & 263 & 0/5 \\
 & Falcon & $k$-means++ & 0.28 & 0.12 & 0.25 & \textbf{0.070} & \textbf{359} & 2/5 \\
\midrule
 DS6 & TF-IDF & FuzzyCM & 0.51 & 0.19 & 0.20 & 0.01 & 74 & 0/5 \\
 & BERT & Spectral & 0.51 & \textbf{0.32} & 0.35 & 0.02 & 37 & 1/5 \\
 & OpenAI & FuzzyCM & \textbf{0.52} & 0.23 & 0.21 & \textbf{0.10} & \textbf{1095} & 3/5 \\
 & LLaMA-2 & k-means++ & 0.19 & 0.08 & \textbf{0.63} & 0.07 & 518 & 1/5 \\
 & Falcon & FuzzyCM & 0.22 & -0.03 & 0.21 & 0.00 & 930 & 0/5 \\
\bottomrule
\end{tabular}

\end{table}

Results of the summarisation experiment, depicted in Table~\ref{tab:sum-results}, show that using summarisation as a dimensionality reduction technique does not consistently benefit all models. Clustering results for the original texts without generated summaries are generally higher than those with summarisation. This finding suggests that essential details necessary for accurate clustering might have been lost during the summarisation process. Alternatively, the inherent complexity and variations of textual representation might require a more sophisticated approach to  text summarisation that can maintain essential information while reducing complexity. Additionally, it is important to highlight that we observed low-quality clustering results when using the summarisation output from the smaller-sized LLaMA-2-7b and Falcon-7b models, likely due to their limited ability to capture and reproduce the complexity and subtle gradations in the source texts.

Results for the model size experiment, presented in Table~\ref{tab:size-results}, highlight the negative influence of the number of parameters on clustering outcomes for Falcon, given that the larger model, Falcon-40b, produced lower ARI, $F1$-score, and HS for the CSTR and SyskillWebert datasets. This may be because embeddings for the Falcon-40b model were created over a subset of the data used for the Falcon-7b embeddings (1.5 trillion tokens for Falcon-7b and 1 trillion tokens for Falcon-40b). In the case of LLaMA-2, models with sizes 7b and 13b were reportedly trained on identical datasets~\cite{Llama27bchathf}. Results indicate that embeddings from the larger model, LLaMA-2-13b, outperform those from LLaMA-2-7b. This outcome is anticipated, as larger models with more parameters generally have a greater capacity to capture complex patterns and relationships in the data, leading to richer and more expressive embeddings.

\begin{figure}[!ht]
	\centering
	\subfloat[LLaMA-2-7b-chat-hf.\label{fig:emb_vis:l7b}]{%
	\includegraphics[width=0.45\linewidth]{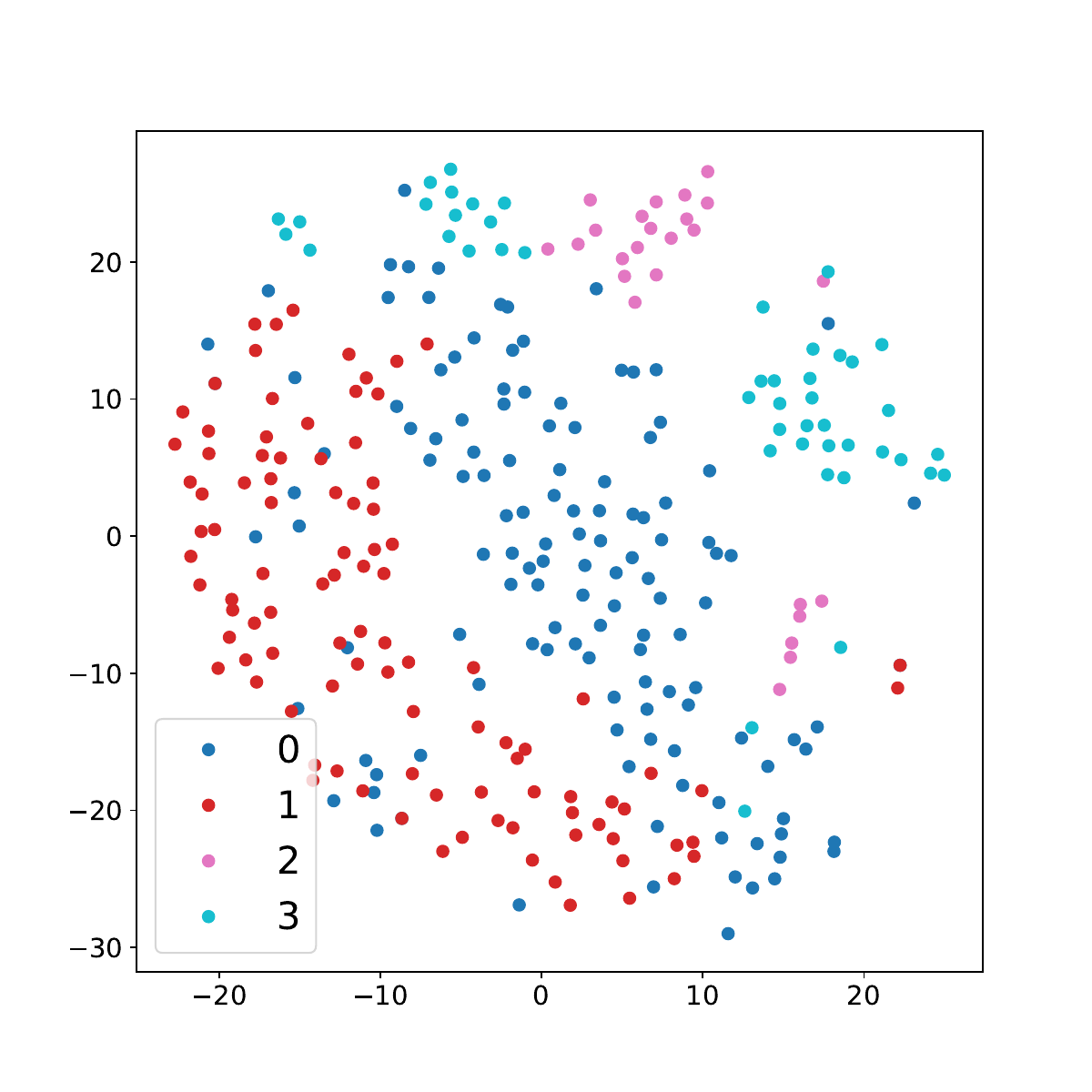}}
    \hfil
    \subfloat[LLaMA-2-13b-chat-hf.\label{fig:emb_vis:l13b}]{%
    	\includegraphics[width=0.45\linewidth]{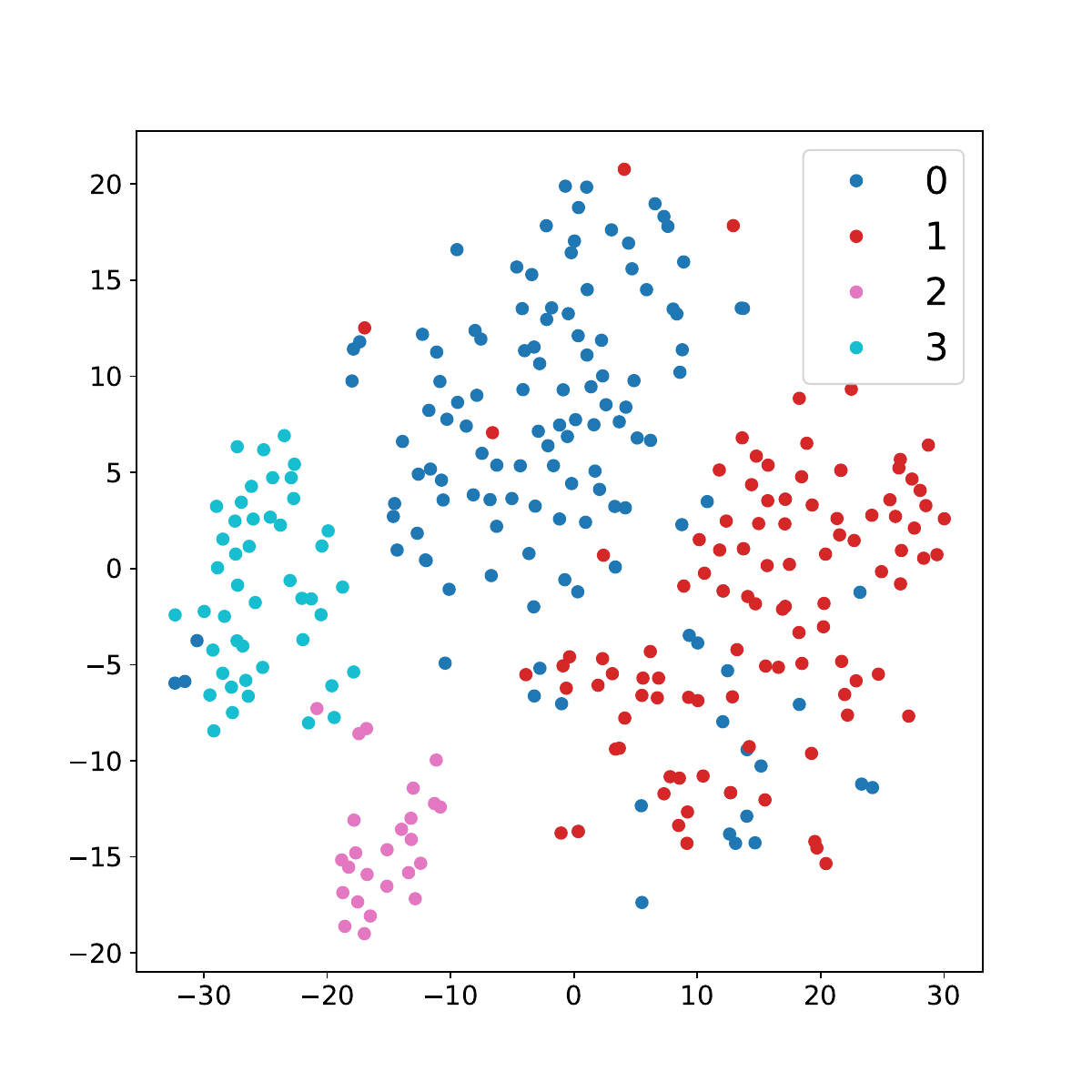}}
      \\
    \subfloat[Falcon-7b.\label{fig:emb_vis:f7b}]{%
    	\includegraphics[width=0.45\linewidth]{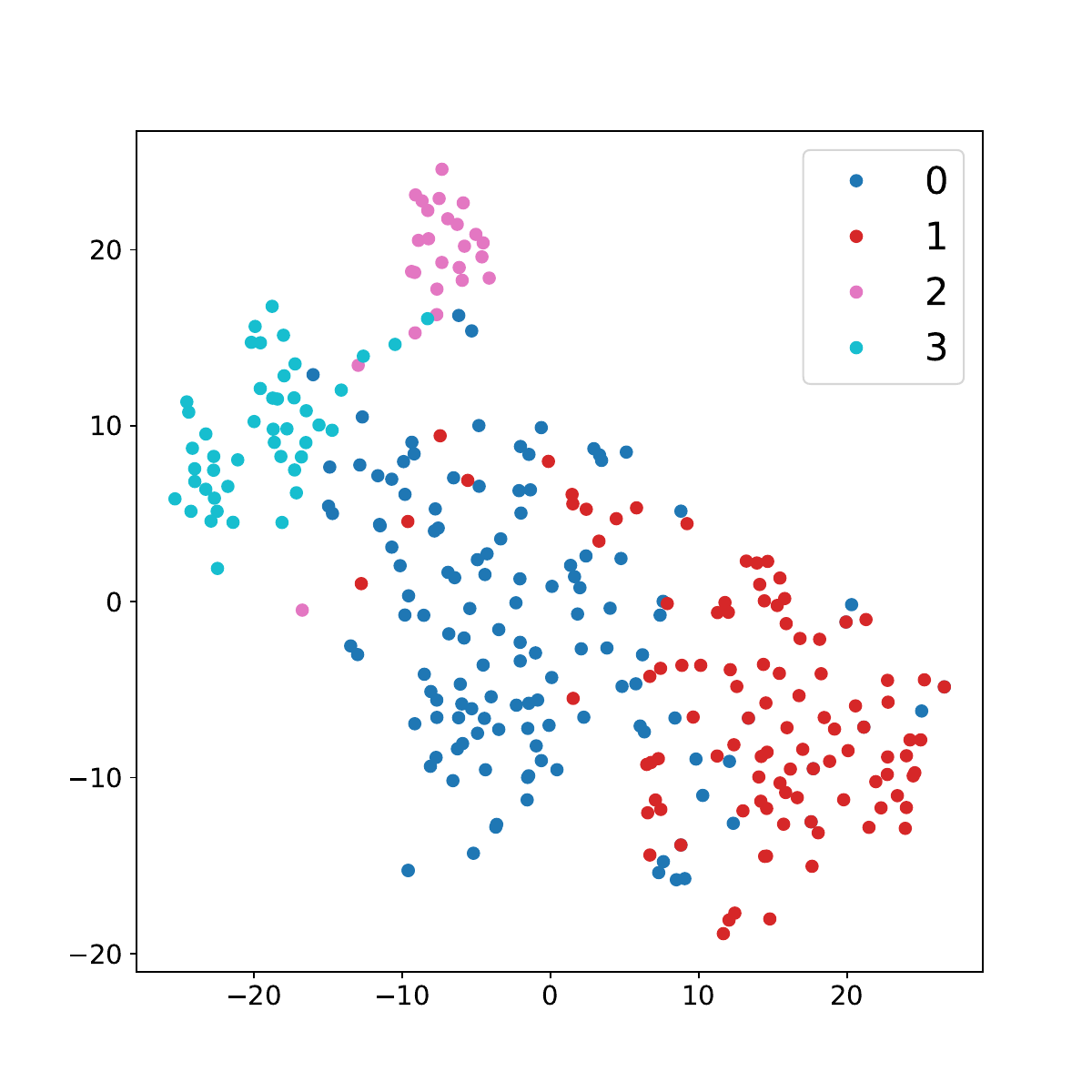}}
     \hfil
     \subfloat[Falcon-40b.\label{fig:emb_vis:f40b}]{%
    	\includegraphics[width=0.45\linewidth]{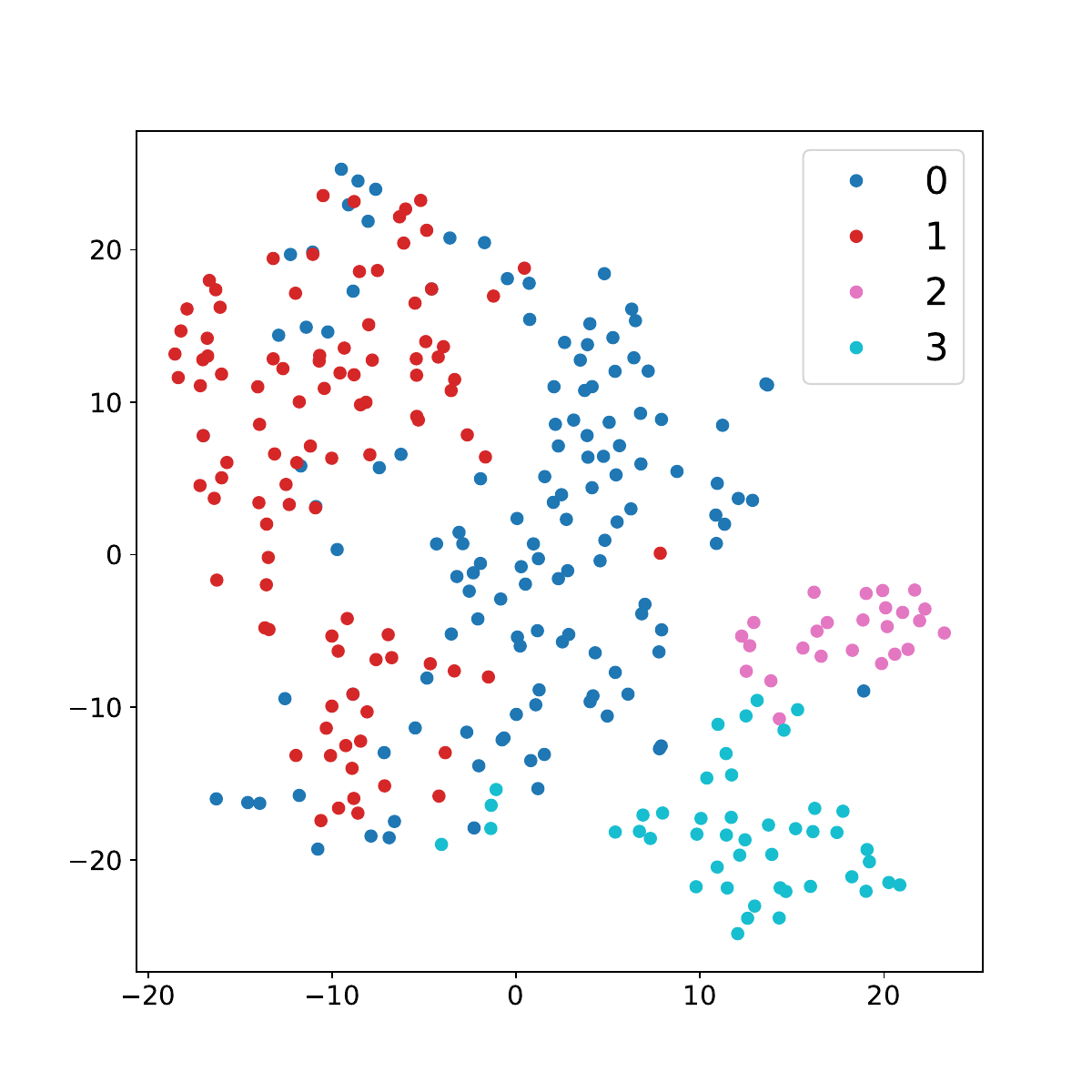}}
    \caption{Representation of different embeddings for the CSTR dataset, where PCA was used as a preliminary dimensionality reduction algorithm and t-SNE for data projection into a lower-dimensional space.}
	\label{fig:emb_vis}
\end{figure}

Another perspective is given by Figure~\ref{fig:emb_vis}, which shows a visualisation of different LLM parameter sizes for the CSTR dataset using PCA and $t$-SNE. A noticeable artefact for LLaMA-2-7b (Figure~\ref{fig:emb_vis:l7b}) and Falcon-40b (Figure~\ref{fig:emb_vis:f40b}) is the lack of coherence in the four document classes, indicating an unclear delimitation in the feature space. This observation is consistent with the results in Table~\ref{tab:size-results}, where these two models have the lowest F1S, ARI, and HS values. Conversely, when employing LLaMA-2-13b (Figure~\ref{fig:emb_vis:l13b}) and Falcon-7b (Figure~\ref{fig:emb_vis:f7b}), the classes exhibit better separation, which aligns with the clustering results observed in Table~\ref{tab:size-results}, where these two models achieved the highest values in 3 out of the 5 metrics calculated. Increasing the number of tokens in Falcon yielded better results, aligning with existing literature that suggests models trained on larger and more diverse corpora are more capable of capturing complex patterns in the data~\cite{Naveed2023ACO}.

Balancing computational requirements with clustering quality involves weighing the benefits of larger embeddings against their resource demands. As results show, a larger number of parameters do not necessarily lead to improved clustering and require significantly more computational power and memory. Empirical evaluation is essential to determine if the improvements justify the additional costs, considering the specific task and resource constraints.

\begin{table}[H]
\caption{Results of summarisation effect on text clustering for the best-performing clustering algorithms in comparison to clustering without summarisation. The `DS' column indicates the dataset being analysed; in particular, DS1 represents the CSTR dataset, while DS2 denotes the SyskillWebert dataset.}
\label{tab:sum-results}

\begin{tabular}{p{0.25in}p{0.75in}p{0.6in}p{0.85in}wr{0.6cm}wr{0.7cm}wr{0.7cm}wr{0.9cm}wr{0.65cm}}
\toprule
DS & Embed. & Version & Best Alg. & F1S & ARI & HS & SS & CHI \\
\midrule
DS1 & BERT & Full & Spectral & \textbf{0.85} & \textbf{0.60} & \textbf{0.63} & \textbf{0.118} & \textbf{25} \\
 &  & Summary & Spectral & 0.81 & 0.50 & 0.56 & 0.114 & 24 \\
 \cmidrule(lr){2-9}
 & OpenAI & Full & $k$-means & \textbf{0.84} & \textbf{0.59} & \textbf{0.64} & \textbf{0.066} & 13 \\
 & & Summary & $k$-means & 0.81 & 0.53 & 0.58 & 0.061 & 13 \\
 \cmidrule(lr){2-9}
 & LLaMA-2 & Full & $k$-means & 0.44 & 0.12 & 0.21 & \textbf{0.099} & \textbf{53} \\
 & & Summary & AHC & \textbf{0.47} & \textbf{0.16} & \textbf{0.30} & 0.072 & 24 \\
 \cmidrule(lr){2-9}
 & Falcon & Full & $k$-means & \textbf{0.74} & \textbf{0.39} & \textbf{0.48} & 0.111 & 34 \\
 & & Summary & $k$-means & 0.40 & 0.03 & 0.02 & \textbf{0.224} & \textbf{329} \\
\midrule
DS2 & BERT & Full & AHC & 0.74 & \textbf{0.58} & 0.53 & \textbf{0.152} & \textbf{37} \\
 & & Summary & AHC & \textbf{0.75} & 0.57 & \textbf{0.54} & 0.089 & 22 \\
 \cmidrule(lr){2-9}
 & OpenAI & Full & AHC & \textbf{0.90} & \textbf{0.79} & \textbf{0.75} & \textbf{0.070} & \textbf{19} \\
 & & Summary & Spectral & 0.79 & 0.71 & 0.64 & 0.054 & 18 \\
 \cmidrule(lr){2-9}
 & LLaMA-2 & Full & $k$-means & \textbf{0.51} & \textbf{0.21} & \textbf{0.25} & 0.137 & 69 \\
 & & Summary & FuzzyCM & 0.25 & 0.04 & 0.06 & \textbf{0.548} & \textbf{603} \\
 \cmidrule(lr){2-9}
 & Falcon & Full & $k$-means++ & \textbf{0.45} & \textbf{0.26} & \textbf{0.30} & 0.170 & 85 \\
 & & Summary & FuzzyCM & 0.34 & 0.04 & 0.07 & \textbf{0.269} & \textbf{577} \\
\bottomrule
\end{tabular}

\end{table}

\begin{table}[!ht]
\caption{Results for the model size experiment, displaying the best performing clustering algorithm comparing two sizes of the LLaMA and Falcon models on the DS1 (CSTR) and DS2 (SyskillWebert) datasets. Model sizes are given in billions of parameters.}
\label{tab:size-results}

\begin{tabular}{p{0.45in}p{1in}p{0.85in}wr{0.6cm}wr{0.7cm}wr{0.7cm}wr{0.9cm}wr{0.65cm}}
\toprule
Dataset & Embed. & Best Alg. & F1S & ARI & HS & SS & CHI \\
\midrule
DS1 & LLaMA-2-7b & $k$-means & 0.41 & 0.09  & 0.17 & \textbf{0.112} & \textbf{50} \\
 & LLaMA-2-13b & AHC & \textbf{0.82} & \textbf{0.53} & \textbf{0.61} & 0.084  & 21 \\
 & Falcon-7b & $k$-means & \textbf{0.74} & \textbf{0.39} & \textbf{0.48} & 0.111  & 34 \\
 & Falcon-40b & AHC & 0.46 & 0.17 & 0.30 & 0.111 & \textbf{44} \\
\midrule
DS2 & LLaMA-2-7b & $k$-means & 0.51 & 0.21 & 0.25 & \textbf{0.137} & \textbf{69} \\
 & LLaMA-2-13b & $k$-means++ & \textbf{0.60} & \textbf{0.49} & \textbf{0.39} & 0.095  & 37 \\
 & Falcon-7b & $k$-means++ & \textbf{0.45} & \textbf{0.26} & \textbf{0.30} & 0.170 & 85 \\
 & Falcon-40b & $k$-means++ & 0.40 & 0.15 & 0.13 & \textbf{0.188} & \textbf{131} \\
\bottomrule
\end{tabular}

\end{table}

In summary, these results provide valuable insights into the relationship between text embeddings, clustering algorithms, and dimensionality reduction techniques. We evaluated five embeddings and four clustering algorithms across five datasets, including the MN-DS dataset with hierarchical labels. Embeddings from the OpenAI model generally outperformed traditional techniques such as BERT and TF-IDF, aligning with previous studies. Conversely, LLaMA-2 and Falcon models produced inferior results compared to BERT across most metrics. The $k$-means algorithm was the most effective on most datasets, while FuzzyCM performed better on the Reuters dataset. Additionally, summarisation models used for dimensionality reduction did not enhance clustering performance, and the use of higher-dimensional models displayed mixed results.

\section{Limitations}
\label{sec:limitations}

Despite the insights revealed in this study, the available computational resources constrained the experiments. This limitation prohibited additional trials with summary generation, higher-dimensional models on more voluminous datasets, and fine-tuning of model parameters. Consequently, the potential benefits of summaries in large-scale text clustering have yet to be thoroughly evaluated. Summarisation might behave differently when applied at scale or with distinct prompts, potentially providing more pronounced benefits or drawbacks depending on the complexity and diversity of the text data.

Moreover, embeddings computed for extra-large models like Falcon-180b or LLaMA-2-70b could potentially deliver substantial gains. However, practical application is restrained by the significant computational demands associated with larger model sizes. This limitation potentially skews the understanding of the absolute efficacy of the embeddings computed for these models, as performance improvements could only be inferred up to a specific size.

Testing the selected clustering algorithms with a wider range of parameters could also provide additional insights into the results. However, due to the aforementioned computational limitations, a breadth-first approach was chosen---testing several different clustering algorithms---rather than a depth-first approach, which would involve experimenting with more parameters for each algorithm. For example, in AHC, the analysis was limited to using Euclidean distance with Ward linkage. Ward linkage was chosen because of its ability to minimize the variance within clusters, resulting in more compact and cohesive clusters, which is particularly beneficial for text data clustering. While this approach provides a comprehensive comparison across different methods, future research could explore parameter optimization in greater depth to further refine clustering outcomes.

These constraints prompt essential considerations for future research. Specifically, experiments with larger datasets and higher-dimension models would enable a more comprehensive and accurate understanding of the potentials and limitations of text clustering algorithms and their scalability in real-world applications.

\section{Conclusions}
\label{sec:conclusion}

In this study, we examined the impact that various embeddings---namely TF-IDF, BERT, OpenAI, LLaMA-2, and Falcon---and clustering algorithms have on grouping textual data. Through detailed exploration, we evaluated the efficacy of dimensionality reduction via summarisation and the role of model size on the clustering efficiency of various datasets. We found that OpenAI's embeddings generally outperform other embeddings, with BERT's performance excelling among open-source alternatives, underscoring the potential of advanced models to positively influence text clustering results.

A key finding from the experiments with summarisation is that summary-based dimensionality reduction does not consistently improve clustering performance. This indicates that careful consideration is required when preprocessing text to avoid losing essential information.

This research highlights the trade-off between improved clustering performance and the computational costs of using embeddings from larger models. Although results indicate that an increase in model size may yield superior clustering performance, potential benefits must be weighed against the practicality of available computing resources.

These findings point towards continued research focused on developing strategies that leverage the strengths of advanced models while mitigating their computational demands. It is also critical to expand the scope of research to include more diverse text types, which will provide a more comprehensive understanding of clustering dynamics across different contexts. Analyzing embeddings of very recent or yet-to-be-released models will also be important. Additionally, conducting parameter searches for clustering algorithms is essential to optimize their performance and ensure robust clustering outcomes across various datasets.

In conclusion, our findings highlight the relationship between embedding types, dimensionality reduction, model size, and text clustering effectiveness in the context of structured, formal texts. While more advanced embeddings such as those from OpenAI offer clear advantages, researchers and practitioners must weigh the trade-offs regarding cost, computational resources, and the effects of text preprocessing techniques.

\section*{Acknowledgements}

This work was financed by the Fundação para a Ciência e a Tecnologia, in the framework of projects UIDB\allowbreak/04111\allowbreak/2020, UIDB\allowbreak/00066\allowbreak/2020, CEECINST\allowbreak/00002\allowbreak/2021\allowbreak/CP2788\allowbreak/CT0001, CEECINST\allowbreak/00147\allowbreak/2018\allowbreak/CP1498\allowbreak/CT0015 as well as Instituto Lus\'ofono de Investiga\c{c}\~ao e Desenvolvimento (ILIND) under project COFAC\allowbreak/ILIND\allowbreak/COPELABS\allowbreak/1\allowbreak/2022.

 \bibliographystyle{elsarticle-num}

\begin{thebibliography}{10}
\expandafter\ifx\csname url\endcsname\relax
  \def\url#1{\texttt{#1}}\fi
\expandafter\ifx\csname urlprefix\endcsname\relax\def\urlprefix{URL }\fi
\expandafter\ifx\csname href\endcsname\relax
  \def\href#1#2{#2} \def\path#1{#1}\fi

\bibitem{Salton1983}
G.~Salton, M.~J. McGill, Introduction to Modern Information Retrieval, McGraw-Hill, 1983.

\bibitem{Ramos2003}
J.~Ramos, Using {TF-IDF} to determine word relevance in document queries, in: Proceedings of the First Instructional Conference on Machine Learning (ICML 2003), 2003, p. 29–48.

\bibitem{Mikolov2013}
T.~Mikolov, K.~Chen, G.~S. Corrado, J.~Dean, \href{https://api.semanticscholar.org/CorpusID:5959482}{{Efficient Estimation of Word Representations in Vector Space}}, in: International Conference on Learning Representations, 2013, pp. 1--12.
\newline\urlprefix\url{https://api.semanticscholar.org/CorpusID:5959482}

\bibitem{Pennington2014}
J.~Pennington, R.~Socher, C.~Manning, {Glove: Global Vectors for Word Representation}, in: Proceedings of the 2014 Conference on Empirical Methods in Natural Language Processing (EMNLP), Vol.~14, Association for Computational Linguistics, 2014, pp. 1532--1543.
\newblock \href {https://doi.org/10.3115/v1/D14-1162} {\path{doi:10.3115/v1/D14-1162}}.

\bibitem{Devlin2018}
J.~Devlin, M.-W. Chang, K.~Lee, K.~Toutanova, {BERT:} pre-training of deep bidirectional transformers for language understanding., in: J.~Burstein, C.~Doran, T.~Solorio (Eds.), NAACL-HLT (1), Association for Computational Linguistics, 2019, pp. 4171--4186.

\bibitem{Brown2020}
T.~Brown, B.~Mann, N.~Ryder, M.~Subbiah, J.~D. Kaplan, P.~Dhariwal, A.~Neelakantan, P.~Shyam, G.~Sastry, A.~Askell, S.~Agarwal, A.~Herbert-Voss, G.~Krueger, T.~Henighan, R.~Child, A.~Ramesh, D.~Ziegler, J.~Wu, C.~Winter, C.~Hesse, M.~Chen, E.~Sigler, M.~Litwin, S.~Gray, B.~Chess, J.~Clark, C.~Berner, S.~McCandlish, A.~Radford, I.~Sutskever, D.~Amodei, {Language Models are Few-Shot Learners}, Advances in Neural Information Processing Systems 33 (2020) 1877--1901.

\bibitem{MacQueen1967}
J.~MacQueen, Some methods for classification and analysis of multivariate observations, in: Proceedings of the Fifth Berkeley Symposium on Mathematical Statistics and Probability, Vol.~1, University of California Press, Berkeley, Calif, 1967, pp. 281--297.

\bibitem{Ward1963}
J.~H. Ward, Hierarchical grouping to optimize an objective function, Journal of the American Statistical Association 58~(301) (1963) 236--244.

\bibitem{Ng2002}
A.~Y. Ng, M.~I. Jordan, Y.~Weiss, On spectral clustering: Analysis and an algorithm, in: Advances in Neural Information Processing Systems, Vol.~2, MIT Press, Cambridge, MA, USA, 2002, pp. 849--856.

\bibitem{Bezdek1981}
J.~C. Bezdek, Pattern Recognition with Fuzzy Objective Function Algorithms, Plenum Press, New York, 1981.

\bibitem{Xie2016}
J.~Xie, R.~Girshick, A.~Farhadi, Unsupervised deep embedding for clustering analysis, in: International Conference on Machine Learning, JMLR.org, 2016, pp. 478--487.

\bibitem{KAMAL2022}
K.~Berahmand, F.~Daneshfar, A.~Golzari~Oskouei, M.~Dorosti, M.~Aghajani, An improved deep text clustering via local manifold of an autoencoder embedding, SSRN Electronic Journal (01 2022).
\newblock \href {https://doi.org/10.2139/ssrn.4295242} {\path{doi:10.2139/ssrn.4295242}}.

\bibitem{Strehl2002}
A.~Strehl, J.~Ghosh, Cluster ensembles -- a knowledge reuse framework for combining multiple partitions, Journal of Machine Learning Research 3 (2002) 583--617.

\bibitem{Pugachev2021}
L.~Pugachev, M.~Burtsev, Short text clustering with transformers, in: International Conference in Computational Linguistics and intelligent technologies, CEUR-WS.org, 2021, pp. 571--577.
\newblock \href {https://doi.org/10.28995/2075-7182-2021-20-571-577} {\path{doi:10.28995/2075-7182-2021-20-571-577}}.

\bibitem{zhang2019integrating}
J.~Zhang, P.~Lertvittayakumjorn, Y.~Guo, Integrating semantic knowledge to tackle zero-shot text classification, in: J.~Burstein, C.~Doran, T.~Solorio (Eds.), Proceedings of the 2019 Conference of the North {A}merican Chapter of the Association for Computational Linguistics: Human Language Technologies, Volume 1 (Long and Short Papers), Association for Computational Linguistics, Minneapolis, Minnesota, 2019, pp. 1031--1040.
\newblock \href {https://doi.org/10.18653/v1/N19-1108} {\path{doi:10.18653/v1/N19-1108}}.

\bibitem{Keraghel2024}
I.~Keraghel, S.~Morbieu, M.~Nadif, Beyond Words: A Comparative Analysis of LLM Embeddings for Effective Clustering, Springer Nature Switzerland, 2024, Ch.~1, pp. 205--216.
\newblock \href {https://doi.org/10.1007/978-3-031-58547-0_17} {\path{doi:10.1007/978-3-031-58547-0_17}}.

\bibitem{yamagishi2019vctk}
J.~Yamagishi, C.~Veaux, K.~MacDonald, {CSTR VCTK Corpus}: English multi-speaker corpus for {CSTR} voice cloning toolkit (version 0.92), online; accessed 28 May 2024 (2019).
\newblock \href {https://doi.org/10.7488/ds/2645} {\path{doi:10.7488/ds/2645}}.

\bibitem{SyskillWebert_dataset}
M.~Pazzani, {SyskillWebert Web Page Ratings}, \url{https://kdd.ics.uci.edu/databases/SyskillWebert/}, online; accessed 28 May 2024 (1999).

\bibitem{Newsgroups20}
T.~Mitchell, Twenty newsgroups, UCI Machine Learning Repository, online; accessed 28 May 2024 (1999).
\newblock \href {https://doi.org/10.24432/C5C323} {\path{doi:10.24432/C5C323}}.

\bibitem{data8050074}
A.~Petukhova, N.~Fachada, {MN-DS}: A multilabeled news dataset for news articles hierarchical classification, Data 8~(5) (2023).
\newblock \href {https://doi.org/10.3390/data8050074} {\path{doi:10.3390/data8050074}}.

\bibitem{lewis2004reuters}
D.~Lewis, Reuters-21578 text categorization collection, online; accessed 28 May 2024 (1997).
\newblock \href {https://doi.org/10.24432/C52G6M} {\path{doi:10.24432/C52G6M}}.

\bibitem{textcl}
A.~Petukhova, N.~Fachada, {TextCL}: A {P}ython package for {NLP} preprocessing tasks, SoftwareX 19 (2022) 101122, online; accessed 28 May 2024.
\newblock \href {https://doi.org/10.1016/j.softx.2022.101122} {\path{doi:10.1016/j.softx.2022.101122}}.

\bibitem{linkTFIDF}
W.~Uther, D.~Mladenić, M.~Ciaramita, B.~Berendt, A.~Kołcz, M.~Grobelnik, M.~Witbrock, J.~Risch, S.~Bohn, S.~Poteet, A.~Kao, L.~Quach, J.~Wu, E.~Keogh, R.~Miikkulainen, P.~Flener, U.~Schmid, F.~Zheng, G.~Webb, S.~Nijssen, {TF–IDF}, Springer US, 2010, Ch. TF–IDF, pp. 986--987.
\newblock \href {https://doi.org/10.1007/978-0-387-30164-8_832} {\path{doi:10.1007/978-0-387-30164-8_832}}.

\bibitem{Zhu_2015_ICCV}
Y.~Zhu, R.~Kiros, R.~S. Zemel, R.~Salakhutdinov, R.~Urtasun, A.~Torralba, S.~Fidler, Aligning books and movies: Towards story-like visual explanations by watching movies and reading books, 2015 IEEE International Conference on Computer Vision (ICCV) (2015) 19--27\href {https://doi.org/10.1109/ICCV.2015.11} {\path{doi:10.1109/ICCV.2015.11}}.

\bibitem{przybyla_2022_6539054}
P.~Przybyła, P.~Borkowski, K.~Kaczyński, Wikipedia complete citation corpus, online; accessed 28 May 2024 (Jul 2022).
\newblock \href {https://doi.org/10.5281/zenodo.6539054} {\path{doi:10.5281/zenodo.6539054}}.

\bibitem{Greene2023Embedding}
R.~Greene, T.~Sanders, L.~Weng, A.~Neelakantan, New and improved embedding model, \url{https://openai.com/blog/new-and-improved-embedding-model}, online; accessed 28 May 2024 (2023).

\bibitem{almazrouei2023falcon}
E.~Almazrouei, H.~Alobeidli, A.~Alshamsi, A.~Cappelli, R.~Cojocaru, M.~Debbah, Étienne Goffinet, D.~Hesslow, J.~Launay, Q.~Malartic, D.~Mazzotta, B.~Noune, B.~Pannier, G.~Penedo, The {F}alcon series of open language models (2023).
\newblock \href {http://arxiv.org/abs/2311.16867} {\path{arXiv:2311.16867}}.

\bibitem{touvron2023llama}
H.~Touvron, L.~Martin, K.~Stone, P.~Albert, A.~Almahairi, Y.~Babaei, N.~Bashlykov, S.~Batra, P.~Bhargava, S.~Bhosale, D.~Bikel, L.~Blecher, C.~C. Ferrer, M.~Chen, G.~Cucurull, D.~Esiobu, J.~Fernandes, J.~Fu, W.~Fu, B.~Fuller, C.~Gao, V.~Goswami, N.~Goyal, A.~Hartshorn, S.~Hosseini, R.~Hou, H.~Inan, M.~Kardas, V.~Kerkez, M.~Khabsa, I.~Kloumann, A.~Korenev, P.~S. Koura, M.-A. Lachaux, T.~Lavril, J.~Lee, D.~Liskovich, Y.~Lu, Y.~Mao, X.~Martinet, T.~Mihaylov, P.~Mishra, I.~Molybog, Y.~Nie, A.~Poulton, J.~Reizenstein, R.~Rungta, K.~Saladi, A.~Schelten, R.~Silva, E.~M. Smith, R.~Subramanian, X.~E. Tan, B.~Tang, R.~Taylor, A.~Williams, J.~X. Kuan, P.~Xu, Z.~Yan, I.~Zarov, Y.~Zhang, A.~Fan, M.~Kambadur, S.~Narang, A.~Rodriguez, R.~Stojnic, S.~Edunov, T.~Scialom, Llama 2: Open foundation and fine-tuned chat models (2023).
\newblock \href {http://arxiv.org/abs/2307.09288} {\path{arXiv:2307.09288}}, \href {https://doi.org/10.48550/arXiv.2307.09288} {\path{doi:10.48550/arXiv.2307.09288}}.

\bibitem{huggingface}
Hugging face, \url{https://huggingface.co/}, online; accessed 28 May 2024 (2024).

\bibitem{arthur2007k}
D.~Arthur, S.~Vassilvitskii, et~al., k-means++: The advantages of careful seeding, in: Proceedings of the 18th Annual ACM-SIAM Symposium on Discrete Algorithms, Vol.~7 of SODA '07, SIAM, 2007, pp. 1027--1035.

\bibitem{scikit-learn}
F.~Pedregosa, G.~Varoquaux, A.~Gramfort, V.~Michel, B.~Thirion, O.~Grisel, M.~Blondel, P.~Prettenhofer, R.~Weiss, V.~Dubourg, J.~Vanderplas, A.~Passos, D.~Cournapeau, M.~Brucher, M.~Perrot, E.~Duchesnay, Scikit-learn: Machine learning in {P}ython, Journal of Machine Learning Research 12 (2011) 2825--2830.

\bibitem{sfuzzy}
J.~Warner, J.~Sexauer, scikit fuzzy, twmeggs, alexsavio, A.~Unnikrishnan, G.~Castelão, F.~A. Pontes, T.~Uelwer, pd2f, laurazh, F.~Batista, alexbuy, W.~V. den Broeck, W.~Song, T.~G. Badger, R.~A.~M. Pérez, J.~F. Power, H.~Mishra, G.~O. Trullols, A.~Hörteborn, 99991, Jdwarner/scikit-fuzzy: Scikit-fuzzy version 0.4.2, online; accessed 28 May 2024 (Nov 2019).
\newblock \href {https://doi.org/10.5281/zenodo.3541386} {\path{doi:10.5281/zenodo.3541386}}.

\bibitem{evaluationmetrics}
N.~Chinchor, \href{https://doi.org/10.3115/1072064.1072067}{{MUC-4} evaluation metrics}, in: Proceedings of the 4th Conference on Message Understanding, MUC4 '92, Association for Computational Linguistics, USA, 1992, p. 22–29.
\newblock \href {https://doi.org/10.3115/1072064.1072067} {\path{doi:10.3115/1072064.1072067}}.
\newline\urlprefix\url{https://doi.org/10.3115/1072064.1072067}

\bibitem{SteinleyDouglas2004}
D.~Steinley, Properties of the hubert-arabie adjusted rand index, Psychological methods 9 (2004) 386--396.
\newblock \href {https://doi.org/10.1037/1082-989X.9.3.386} {\path{doi:10.1037/1082-989X.9.3.386}}.

\bibitem{Rosenberg2007}
A.~Rosenberg, J.~Hirschberg, \href{https://aclanthology.org/D07-1043}{{V}-measure: A conditional entropy-based external cluster evaluation measure}, in: J.~Eisner (Ed.), Proceedings of the 2007 Joint Conference on Empirical Methods in Natural Language Processing and Computational Natural Language Learning ({EMNLP}-{C}o{NLL}), Association for Computational Linguistics, Prague, Czech Republic, 2007, pp. 410--420.
\newline\urlprefix\url{https://aclanthology.org/D07-1043}

\bibitem{Rousseeuw1987}
P.~Rousseeuw, Silhouettes: A graphical aid to the interpretation and validation of cluster analysis, Journal of Computational and Applied Mathematics 20 (1987) 53--65.
\newblock \href {https://doi.org/10.1016/0377-0427(87)90125-7} {\path{doi:10.1016/0377-0427(87)90125-7}}.

\bibitem{Caliński1974}
T.~Caliński, H.~JA, A dendrite method for cluster analysis, Communications in Statistics - Theory and Methods 3 (1974) 1--27.
\newblock \href {https://doi.org/10.1080/03610927408827101} {\path{doi:10.1080/03610927408827101}}.

\bibitem{miller2019leveraging}
D.~Miller, Leveraging {BERT} for extractive text summarization on lectures (2019).
\newblock \href {http://arxiv.org/abs/1906.04165} {\path{arXiv:1906.04165}}.

\bibitem{gpt35turbo}
gpt-3-5-turbo, \url{https://platform.openai.com/docs/models/gpt-3-5-turbo/}, online; accessed 28 May 2024 (2024).

\bibitem{falcon7b}
falcon-7b, \url{https://huggingface.co/tiiuae/falcon-7b/}, online; accessed 28 May 2024 (2024).

\bibitem{Llama27bchathf}
Llama-2-7b-chat-hf, \url{https://huggingface.co/meta-llama/Llama-2-7b-chat-hf/}, online; accessed 28 May 2024 (2024).

\bibitem{van2008visualizing}
L.~Van~der Maaten, G.~Hinton, Visualizing data using {$t$-SNE}., Journal of machine learning research 9~(11) (2008).

\bibitem{Naveed2023ACO}
H.~Naveed, A.~U. Khan, S.~Qiu, M.~Saqib, S.~Anwar, M.~Usman, N.~Barnes, A.~S. Mian, \href{https://api.semanticscholar.org/CorpusID:259847443}{A comprehensive overview of large language models}, ArXiv abs/2307.06435 (2023).
\newline\urlprefix\url{https://api.semanticscholar.org/CorpusID:259847443}

\end{thebibliography}

\end{document}